\documentclass[10pt,twocolumn,letterpaper]{article}

\usepackage{multirow}
\usepackage{wacv}
\usepackage{times}
\usepackage{epsfig}
\usepackage{graphicx}
\usepackage{amsmath}
\usepackage{amssymb}
\usepackage{boldline} 
\usepackage{makecell}
\usepackage{ctable} 

\newcommand{\quotes}[1]{``#1''}

\usepackage{bbding}
\usepackage{pifont}
\usepackage{wasysym}
\usepackage{amssymb} 
\usepackage{rotating}
\usepackage{array}
\usepackage{makecell}
\def\wacvPaperID{2077} 

\wacvfinalcopy 

\ifwacvfinal
\def\assignedStartPage{9876} 
\fi

\ifwacvfinal
\else
\fi

\begin{document}

\title{A Quality Aware Sample-to-Sample Comparison for Face Recognition}

\author{Mohammad Saeed Ebrahimi Saadabadi, Sahar Rahimi Malakshan,\\
Ali Zafari, Moktari Mostofa, and Nasser M. Nasrabadi\\
{\tt\small{me00018, sr00033, az00004, mm0251}@mix.wvu.edu, nasser.nasrabadi@mail.wvu.edu}
}

\maketitle

\begin{abstract}
Currently available face datasets mainly consist of a large number of high-quality and a small number of low-quality samples. As a result, a Face Recognition (FR) network fails to learn the distribution of low-quality samples since they are less frequent during training (underrepresented). Moreover, current state-of-the-art FR training paradigms are based on the sample-to-center comparison (i.e., Softmax-based classifier), which results in a lack of uniformity between train and test metrics. This work integrates a quality-aware learning process at the sample level into the classification training paradigm (QAFace). In this regard, Softmax centers are adaptively guided to pay more attention to low-quality samples by using a quality-aware function. Accordingly, QAFace adds a quality-based adjustment to the updating procedure of the Softmax-based classifier to improve the performance on the underrepresented low-quality samples. Our method adaptively finds and assigns more attention to the recognizable low-quality samples in the training datasets. In addition, QAFace ignores the unrecognizable low-quality samples using the feature magnitude as a proxy for quality. As a result, QAFace prevents class centers from getting distracted from the optimal direction. The proposed method is superior to the state-of-the-art algorithms in extensive experimental results on the CFP-FP, LFW, CPLFW, CALFW, AgeDB, IJB-B, and IJB-C datasets.
\end{abstract}

\section{Introduction}
Recent advances in FR performance can be credited to introduction of novel network architectures, large-scale datasets, and new loss functions \cite{liu2019adaptiveface}. Regarding the architecture, ResNet and its variants are mostly used as the backbone for extracting features from the face images \cite{deng2019arcface}. In terms of datasets,  large-scale publicly available training data leads to unprecedented improvement in FR performance \cite{du2020semi}. 
\begin{figure}[t]
\begin{center}
\includegraphics[width=1.0\linewidth]{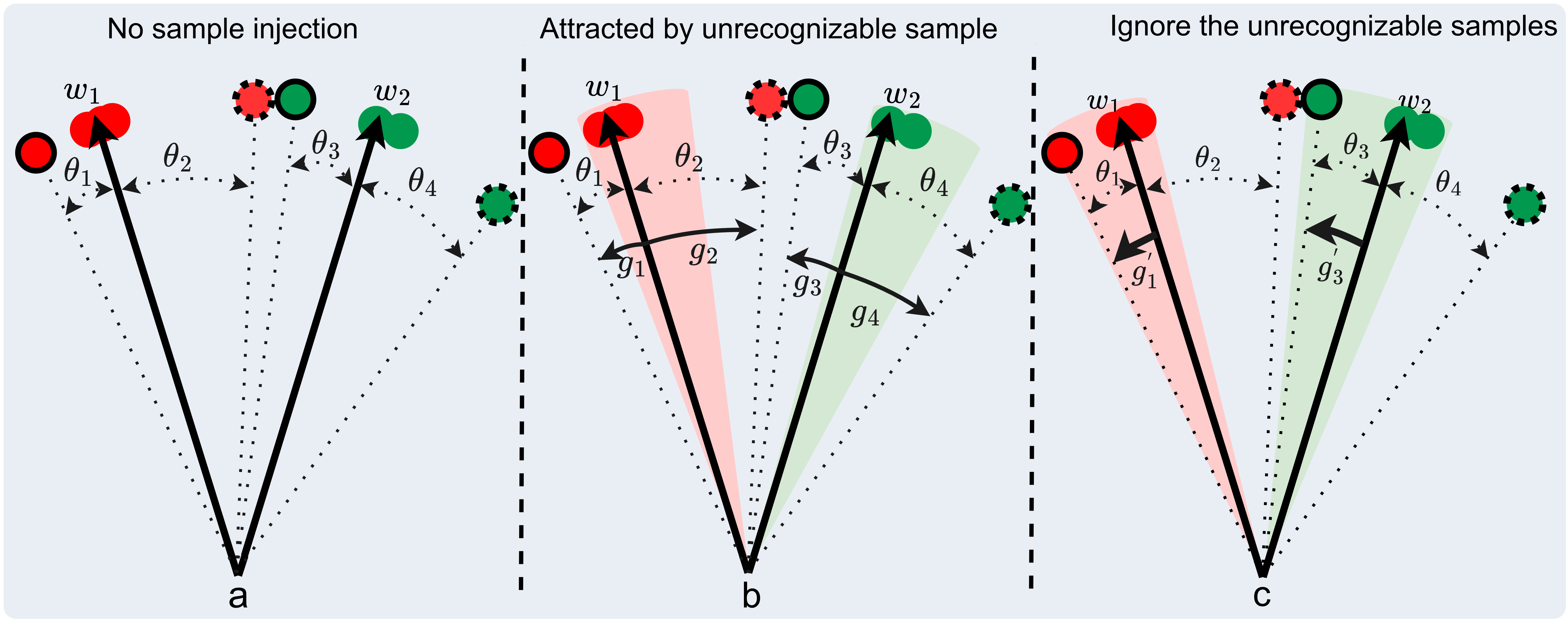}
\end{center}
   \caption{A binary classification example illustrates that unrecognizable samples can misguide the Softmax centers, $w_1$ and $w_2$, from their optimal direction. Circles with solid black borders are recognizable low-quality samples, and black dashed borders are unrecognizable samples. $g_i$ shows the direction that the centers are being pushed. a) without injection, there is no $g$, b) equally injecting samples results in stronger $g$ from samples with more angular disparity, and c) injecting with emphasis on recognizable low-quality samples and ignoring unrecognizable (there is no $g$ toward unrecognizable). Shaded areas are the direction in which feature injection causes the centers to move. Note that $g_1 < g_1^{'}$ and $g_2 < g_2^{'}$.}
\label{fig:lowvshigh}
\vspace{-5mm}
\end{figure}
Recent attempts on FR are mainly focused on manipulating the training criteria \cite{liu2017sphereface,wang2018cosface,deng2019arcface,schroff2015facenet,wang2017normface,wen2021sphereface2}. In this manner, Softmax with a cross-entropy loss, i.e., \ sample-to-center comparison, is the most popular criterion for FR, (i.e., classification) \cite{liu2017sphereface}. In the classification framework, the weights connecting the penultimate layer output (i.e., feature) to the classification layer represent the centers of Softfmax classes  \cite{an2022killing}.

Since FR is an open-set problem,
during testing, sample-to-center (dis)similarity of the Softmax is irrelevant and sample-to-sample (dis)similarity matters.
In order to unify the train and test similarity metrics, pioneering works \cite{schroff2015facenet,wen2016discriminative}, devised metric-based loss functions based on sample-to-sample comparison \cite{sun2014deep,wen2016discriminative}. 
These metric-based losses try to directly minimize a distance metric when two samples come from the same identity (positive pair); otherwise (negative pair) impose a margin \cite{schroff2015facenet}.
However, the sample-wise comparison highly depends on the pair selection strategy and requires a sophisticated mining method \cite{kim2020broadface}. Besides, in large-scale datasets \cite{deng2019arcface} with thousands of identities and millions of samples, there is a combinatorial explosion in the number of possible pairs, leading to costly pair selection,  unstable training, and slow convergence \cite{sun2018multi}.

Several studies show that projecting features and class centers to the unit-hypersphere improves the discriminative power of representation learned by Softmax \cite{wang2018cosface,wang2017normface,liu2017sphereface}. In this manner, Softmax classifies images using the angular distance between the feature representations and Softmax centers \cite{wang2017normface}. An angular margin is then integrated to the Softmax loss to further enforce intra-class compactness and inter-class separability \cite{liu2017sphereface,wang2018cosface,liu2019adaptiveface,zhang2019adacos}. The angular-based Softmax loss functions are more stable than metric-based, i.e., \ no pair selection, and the number of centers is much fewer than the number of samples \cite{liu2022anchorface}. As a result, angular-based Softmax losses have become the state-of-the-art method for training FR frameworks \cite{deng2019arcface,meng2021magface}.

In sample-to-center training paradigms, every identity is represented as a deterministic point in high-dimensional latent space \cite{chang2020data}, i.e., centers. As a result, their performance degrades when there is a large disparity between training and testing data \cite{shi2019probabilistic}.
Although augmentation may narrow the gap between training and evaluation data distribution, it increases the occurrence of unrecognizable samples and overfitting \cite{lv2017data}, as illustrated in Fig. \ref{fig:aug}.
To alleviate this problem, authors in \cite{meng2021magface} propose to use feature magnitude as a proxy to measure the image quality. 
Also, there are methods to estimate the distribution of each class instead of presenting them as a single deterministic point \cite{chang2020data,shi2019probabilistic}. Despite the performance improvement, these approaches do not propose a solution to the overfitting problem nor guarantee (dis)similarity between (negative)positive samples. 

Usually, there are three different types of samples in FR datasets \cite{meng2021magface}. First, samples that are easy to learn by the model. These easy samples usually have good quality-related factors such as high resolution \cite{meng2021magface}. Second, images that have low-quality, but recognizable (hard-samples) \cite{chen2021lightqnet}. Third, unrecognizable samples that even humans can not correctly recognize their identity \cite{meng2021magface,chen2021lightqnet}.
As shown in Fig. \ref{fig:lowvshigh}, unrecognizable samples have larger angular disparity, $\theta_2$ and $\theta_4$, compared to low-quality instances, $\theta_2>\theta_1$ and $\theta_4>\theta_3$. The primary idea in recent works is to increase the margin constraints as the angular disparity between sample and the Softmax center increases \cite{zhang2019adacos,liu2019adaptiveface,huang2020curricularface}.
However, the recognizability of samples and sample-wise (dis)similarity are not considered by any of the mentioned methods. Therefore, the model tries to reduce the training loss by overfitting on unrecognizable samples, which harms model generalization \cite{du2020semi}.

Recently, authors in \cite{deng2021variational} integrate sample-to-sample comparison to the Softmax via injecting sample representations to the centers. 
Although VPL \cite{deng2021variational} brings sample-to-sample comparison to the Softmax framework, the prior assumption is that the unrecognizable samples do not distract the learning. Due to the larger angular disparity, injecting without considering the recognizability of samples puts more emphasis on unrecognizable samples during injection, see Fig. \ref{fig:lowvshigh}.  Adding variations toward the unrecognizable samples harms the model learning paradigm and distract the Softmax centers from the optimal direction. The injection process directly changes centers. Therefore, it is important to push the centers to a valid direction.


Sample selection strategy is indispensable in every sample-wise FR training paradigm \cite{schroff2015facenet,liu2019adaptiveface}. In this work, we try to weigh samples based on their recognizability and quality. In this manner, the sample-wise part of the proposed method (QAFace), injection, benefits from recognizable low-quality (hard) samples. 
During training, QAFace effectively ignores the unrecognizable samples and prevents the class centers from being distracted from the optimal direction. At the same time, QAFace emphasizes on low-quality samples considering them as hard samples. Moreover, since high-quality samples are being well explored by sample-to-center part of the training, the proposed method puts less attention on high-quality
samples during their injection. 
Compared to \cite{deng2021variational}, our method adds no extra memory consumption and sampling strategy to the training.
The presented model queues sample representation using MOCO \cite{he2020momentum} to maintain both samples and centers on the same embedding space.
Contributions of this work can be summarized as follows:
\begin{itemize}
  \item we use informative hard samples (low-quality instances) to introduce sample-wise comparison to the angular-margin Softmax loss.
  \item we propose a new quality-based weighting function that can effectively de-emphasize unrecognizable samples based on the magnitude of their feature representation as a proxy of image quality.
  \item we leverage hard samples to add uncertainty to the Softmax centers toward the direction of hard samples.
\end{itemize}

\section{Related Works}
\subsection{FR Loss Functions} Most of the previous FR methods were established on a metric-learning loss function, such as triplet \cite{schroff2015facenet} or contrastive loss \cite{chopra2005learning,mostofa2022pose}. These loss functions were based on sample-to-sample comparison in Euclidean space. Then \cite{wen2016discriminative} enhanced the intra-class similarity via proposing a new loss to directly minimize intra-class distance while doing classification. 
In this manner, the main challenges toward general FR were the necessity of sample mining, lack of generalization, and feature collapsing problem \cite{wang2018cosface,schroff2015facenet,hashemi2022improving}. More recently, studies showed that applying Softmax to the angular space enhances the discriminability of features \cite{liu2017sphereface,wang2018cosface,deng2019arcface,wen2016discriminative}. Consequently, pioneering works of \cite{wang2017normface,wang2018cosface,deng2019arcface}, introduced intuitive loss functions by applying three different types of margin to the angular space of Softmax: 1) multiplicative angular margin, 2) cosine margin, and 3) additive angular margin, resulting to state-of-the-art performance. 

\subsection{Angular Margin Variations}
Recent studies explore the effects of adaptive angular margin on the learning paradigm of the network \cite{zhang2019adacos,liu2019adaptiveface,liu2019adaptiveface}.
Liu {\it et al.} \cite{liu2019adaptiveface} propose to adaptively tune the margin value to put more constraint on the tail classes. The role of negative samples in obtaining more discriminative features is investigated in \cite{huang2020curricularface}. Authors of \cite{duan2019uniformface}, add new term to the angular-margin loss function to supervise the uniformly spreading of class centers on the unit hyper-sphere.  MagFace \cite{meng2021magface} establishes the norm of features as a proxy of sample recognizability. Image recognizability increases as the feature norm increase \cite{meng2021magface}. MagFace assigns high angular margin on the high-norm feature in the premises of pushing those samples to be closer to their class center. The drawback is that it fails to put emphasis on the valuable hard samples. Furthermore, none of the mentioned methods guarantee the samples-wise similarity. Also, representing each identity with a single deterministic point, i.e.,\ center, in high-dimensional space results in the performance drop when testing data has a large disparity with training samples \cite{shi2019probabilistic}. 

\begin{figure}[t]
\begin{center}
\includegraphics[width=0.9\linewidth]{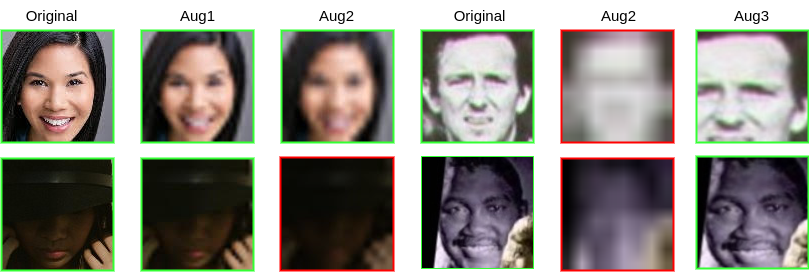}
\end{center}
\vspace{-4mm}
   \caption{Same Augmentation, i.e., down-sampling and random cropping, results in different recognizability among samples. Green border shows the recognizable samples, and red shows unrecognizable samples.}
\label{fig:aug}
\vspace{-5mm}
\end{figure}

\subsection{Probabilistic Face Modeling} 

Probabilistic face modeling is well-established in face template/video matching \cite{cevikalp2010face,arandjelovic2005face}. In these works, a series of samples is used as the input rather than a single face image. Shi {\it et al.} for the first time integrate uncertainty into a single image FR \cite{shi2019probabilistic}. PFE \cite{shi2019probabilistic} represents each image as a Gaussian distribution. The mean and variance of the Gaussian reflect the \quotes{most likely latent feature}, and \quotes{the uncertainty in the feature values}, respectively \cite{shi2019probabilistic}. The goal is to add uncertainty to the model to boost performance for unseen data with large disparity \cite{chang2020data}.
Instead of adding uncertainty to each image representation, VPL \cite{deng2021variational} assigns a distribution to each class within the classification framework. Specifically, VPL injects the class instances to the corresponding classifier to bring more uncertainty to centers and, at the same time, integrate sample-to-sample comparison to the classification paradigm. However, it fails to consider the image recognizability measure. Considering the whole memorizing process in \cite{deng2021variational}, projecting all the representations to the unit hyper-sphere results in an equal contribution of different instances in the memory. Therefore, because of large angular disparity with centers, unrecognizable samples distract centers from their optimal direction, see Fig. \ref{fig:lowvshigh} (b). 
\begin{figure*}
\begin{center}
\includegraphics[width=1.0\linewidth]{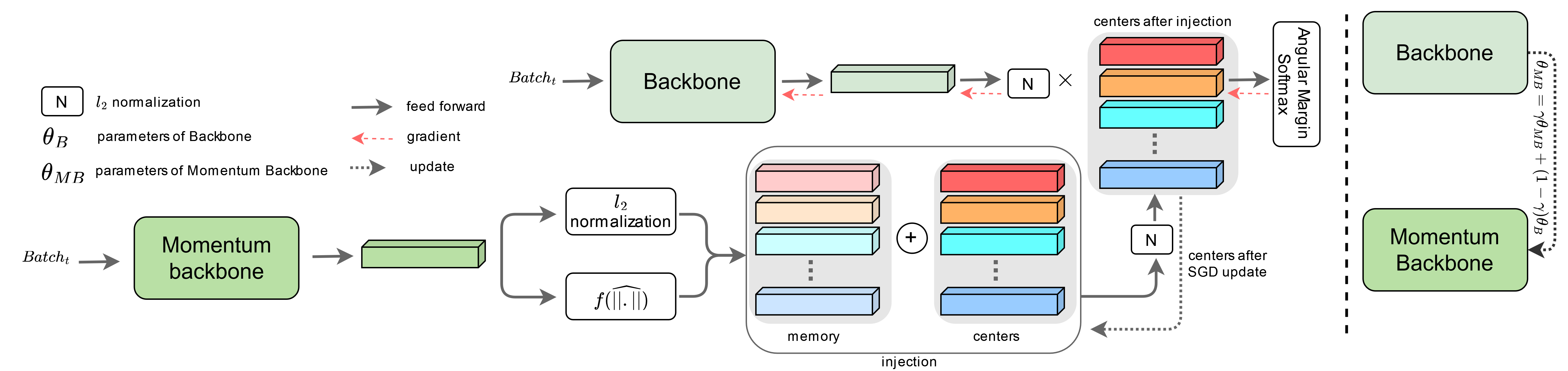}
\end{center}
\vspace{-4mm}
   \caption{Left: the general architecture of the proposed method. In each iteration, the class centers represent the centers and accumulated features of the hard samples of previous iterations. The representations for injecting to the centers are obtained from the momentum backbone. Right: shows the updating of the Momentum Backbone parameters.}
\label{fig:short}
\vspace{-5mm}
\end{figure*}

\section{Proposed Method}
In this section, we begin by analyzing the Softmax-based loss functions. Then, we further explain the integration of Softmax-based classifier with the sample-to-sample comparison. We devise a new injection function to integrate a quality-aware sample-to-sample comparison to the classification framework (QAFace). Finally, we investigate the capability of our method to ignore the unrecognizable samples and the complementary role of our quality-aware injection to the learning signal of Softmax-based loss function.

\subsection{Preliminaries}
Most of the deep visual recognition modules, including FR, can be regarded as the stack of non-linear feature extractor layers (backbone), together with a classifier which is usually a Softmax layer \cite{boutros2021mfr}. Both the backbone and classifier will be trained end-to-end using a back-propagation algorithm.
The Softmax training criterion can be formulated as follows \cite{aghdaie2022morph,kashiani2022robust}:
\begin{equation}\label{Softmax}
 \small
 \begin{aligned}
		L= -\frac{1}{N}\sum_{i=1}^{N}{\log{\frac{e^{W_{y_i}^{T}x_i+b_{y_i}}}{{e^{W_{y_i}^{T}x_i+b_{y_i}}}+{\sum_{\substack{j=1 \\ j \neq y_i}}^{C}{e^{W_{j}^{T}x_i+b_j}}}}}},
\end{aligned}
\end{equation}
where $W_j \in \mathbb{R}^d$ is $j$-th classifier (center), $d$ is the feature dimension, and $b_j$ is the bias for $j$-th Softmax output. $x_i$ is the learned representation of $i$-th sample, and $y_i$ is its corresponding ground truth. $N$ and $C$ represent the mini-batch size and the total number of classes, respectively.

\begin{figure}[t]
\begin{center}
\includegraphics[width=.95\linewidth]{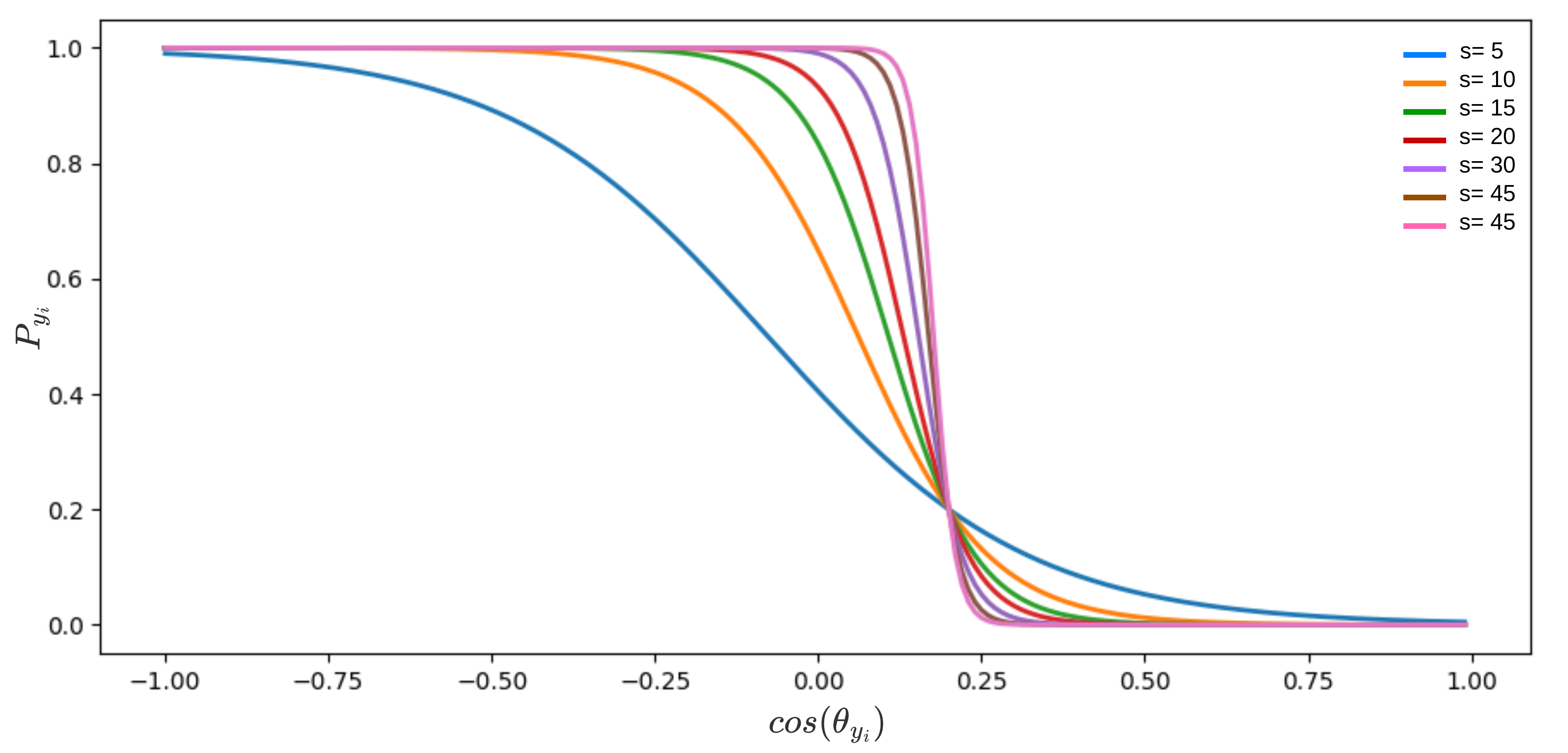}

   \caption{Curve of $\small{p_{i,j} = \frac{e^{W_{j}^{T}x_i}}{{\sum_{{j=1 }}^{C}{e^{W_{j}^{T}x_i}}}}}$ when $\small{cos(\theta_{j})}$ is fixed and $\small{cos(\theta_{y_i})}$ changes from -1.0 to 1.0.}
\label{fig:sAffects}
\vspace{-5mm}
\end{center}
\end{figure}

The angular distribution of representations learned via the Softmax loss, $x_i$, suggests using cosine distance as the metric rather than Euclidean distance \cite{wang2017normface}. Hence, a modified Softmax loss was defined by projecting both centers and representations to the unit-hypersphere \cite{wang2017normface,liu2017sphereface}, $||W_j||=||x_i|| = 1$ and $b_j=0$.
\begin{equation}\label{Method_Sloss}
 \small
 \begin{aligned}
	L^{{'}}= -\frac{1}{N}\sum_{i=1}^{N}{\log{\frac{e^{s  ({cos(\theta_{y_i})})}}{{e^{s  ({cos(\theta_{y_i})})}}+{\sum_{\substack{j=1 \\ j \neq y_i}}^{C}{e^{s  (cos(\theta_j))}}}}}},
\end{aligned}
\end{equation}
where $cos(\theta_{y_i})$ reflects the cosine similarity between $x_i$ and $w_{y_i}$ and $cos(\theta_j)$ denote similarity between $x_i$ and $w_j$ (negative centers).   
$s$ is introduced as the scaling hyper-parameter which affects the curves of the output \cite{zhang2019adacos}, see Fig. \ref{fig:sAffects}. In common FR practice, biases are removed from Eq. \ref{Softmax} because they are learned for close-set recognition and cannot be generalized to open-set testing. 

To enhance the intra-class compactness and inter-class separability, authors in \cite{wang2017normface,wang2018cosface,deng2019arcface} developed intuitive loss functions by applying three different types of margins to Eq. \ref{Method_Sloss}. 
\begin{equation}\label{SCAloss}
 \small
 \begin{aligned}
	L^{{''}}= -\frac{1}{N}\sum_{i=1}^{N}{\log{\frac{e^{{s(cos(m_{S}\theta_{y_i}+m_{A})-m_{\tiny{C}})}}}{{e^{{s(cos(m_{S}\theta_{y_i}+m_{A})-m_{\tiny{C}})}}}+{\sum_{\substack{j=1 \\ j \neq y_i}}^{C}{e^{s cos(\theta_j)}}}}}},
\end{aligned}
\end{equation}

\noindent SphereFace \cite{liu2017sphereface} introduced the multiplicative angular margin to modify decision boundaries from $cos(\theta_1)=cos(\theta_2)$ to $cos(m_S\theta_1)=cos(\theta_2)$. Where ($\theta_1$)$\theta_2$ represent angle between $x_i$ and ($W_{y_i}$)$W_j$. Their modification to Softmax does improve the result; however, the proposed loss function is computed through a series of approximations, which results in unstable training \cite{wang2018cosface}. CosFace modified the decision boundary to $cos(\theta_1)+m_C = cos(\theta_2)$ and ArcFace changed it to $cos(\theta_1 + m_A)=cos(\theta_2)$.  Eq.~\ref{SCAloss} represents all mentioned modifications. Where $m_S$, $m_C$, and $m_A$ are margins introduced by SphereFace \cite{liu2017sphereface}, CosFace \cite{wang2018cosface}, and ArcFace \cite{deng2019arcface}, respectively.
Despite the remarkable improvement, the sample-wise (dis)similarity is not considered in any of these modifications. 

Also, presenting each identity with a single deterministic point in embedding space results in performance degradation when there is a significant disparity between training and testing data \cite{shi2019probabilistic}. For better illustration, we experiment by manually degrading five high-quality testing datasets. Comparing the results of Arcface and VPL in Table \ref{table:lowresperformance}, less performance gap in VPL shows that adding uncertainty to the Softmax centers results in better handling of quality disparity between train and test datasets. Comparing the results of QAFace with VPL, it is shown that our proposed method can further reduce the gap between representation of high and low quality samples by putting more emphasise on the low-quality samples during the injection.
\subsection{Classification-Based Gradient}
We can divide a Softmax-based FR method into its backbone and classifier components.
Hence, here we study the updating of the backbone and center of Softmax separately. For the backbone, we show the gradient with regard to its output (feature), i.e., $\frac{\partial L}{\partial x_i}$.
By omitting bias in the Eq.~\ref{Softmax} the derivatives to $j$-th class center and $i$-th sample's feature are: 
\begin{equation}\label{Method_Xgrad}
 \small
 \begin{aligned}
	\frac{\partial L}{\partial x_i}= ((p_{i,y_i}-1)W_{y_i})+\sum_{\substack{j=1 \\ j \neq y_i}}^{C}{p_{i,j}}W_j,
\end{aligned}
\end{equation}
\begin{equation}\label{Method_Wgrad}
 \small
 \begin{aligned}
	\frac{\partial L}{\partial W_J}= \sum_{\substack{i=1 \\ y_i=j}}^{N}{((p_{i,y_{i}}-1)x_{i})}+\sum_{\substack{i=1 \\ y_i \neq j}}^{N}{p_{i,j}}x_i,
\end{aligned}
\end{equation}
where $\small{p_{i,j} = \frac{e^{W_{j}^{T}x_i}}{{\sum_{\substack{j=1 }}^{C}{e^{W_{j}^{T}x_i}}}}}$. 
Eq.~\ref{Method_Xgrad} shows that from the backbone perspective, the network is being updated toward increasing the similarity between features and the positive class centers while decreasing similarity with negative centers. Moreover, Eq.~\ref{Method_Wgrad} demonstrates that centers update toward being more similar to their corresponding class instances and away from samples of other classes. Hence, both backbone and centers are moving toward each other, and sample-wise (dis)similarity is being supervised indirectly.

\subsection{Sample-Wise Similarity with Softmax}
In order to directly supervise sample-wise (dis)similarity, \cite{deng2021variational} injects samples feature to their corresponding class center.
To this end, a memory, $M$, is constructed, which memorizes the positive features of each class. The memory has the same shape as the Softmax centers: $W \in \mathbb{R}^{C\times d}$, and $M \in \mathbb{R}^{C\times d}$. Considering the injection process as: $\widetilde{W}_{y_i}=W_{y_i}+\lambda M_{y_i}$, the derivative with regard to features changes to:
\begin{equation}\label{Method_WVPLgrad}
 \small
 \begin{aligned}
	\frac{\partial L}{\partial x_i}= ((p_{i,y_i}-1)(W_{y_i}+\lambda M_{y_i}))+\sum_{\substack{j=1 \\ j \neq y_i}}^{C}{p_{i,j}}(W_j+\lambda M_j),
\end{aligned}
\end{equation}
here, the memorized features, $M$, affects the gradient that is updating the backbone. Therefore, sample-to-sample (dis)similarity is being directly supervised.
$\lambda$ is a hyper-parameter to adjust the amount of the injection and should be set manually.
In this injection manner, all the representations are projected to the unit hyper-sphere first. Therefore, unrecognizable samples which have considerable angular disparity with Softmax centers have more influence on the centers than other samples. Therefore, centers will be distracted from the optimal direction. On the other hand, high-quality samples have high similarity with class centers and do not contribute to adding beneficiary variation to the centers.

\begin{figure}
\begin{center}
\includegraphics[width=1.0\linewidth]{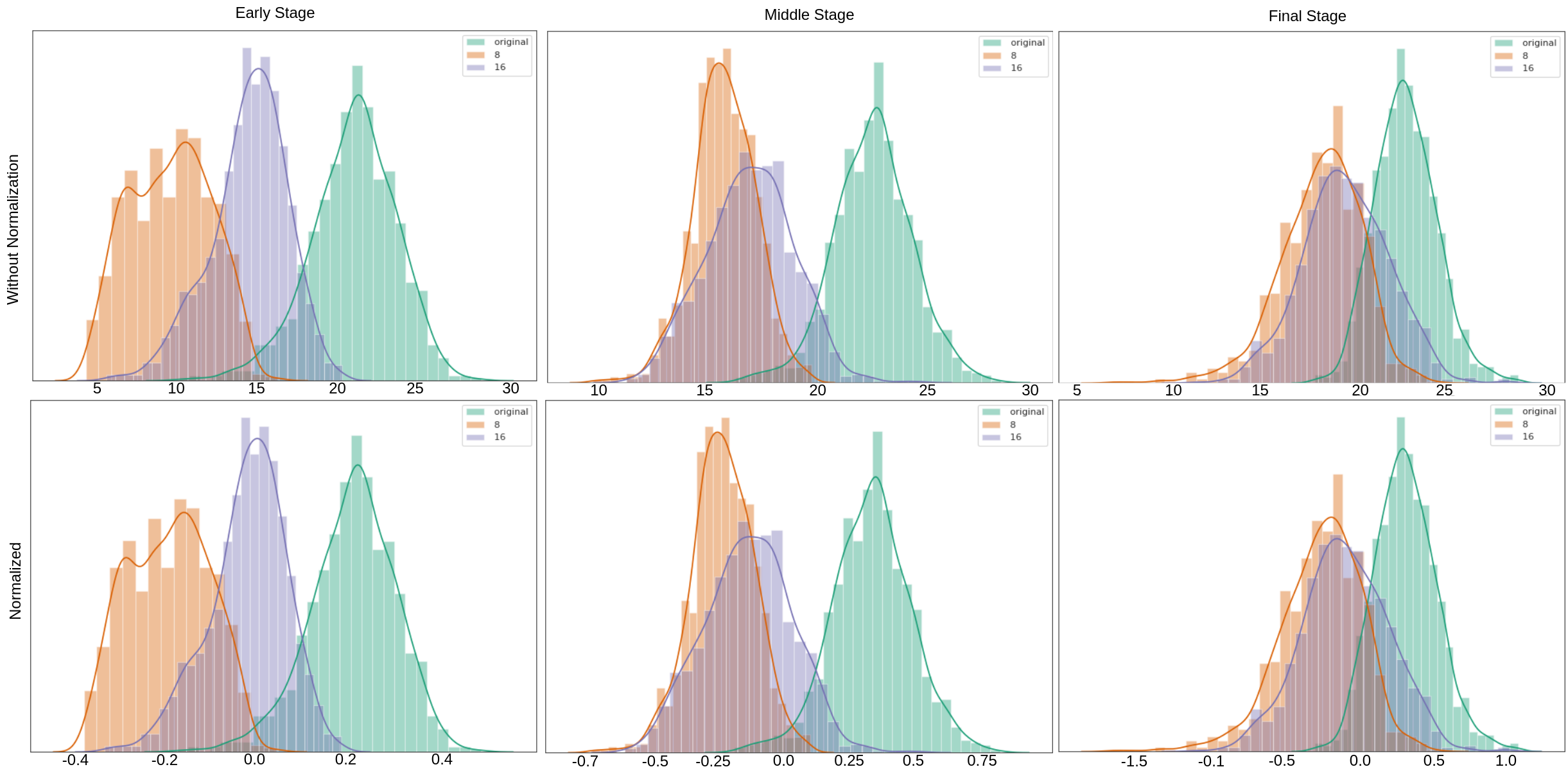}
\end{center}
   \caption{Histogram of the magnitude of features obtained from 10k randomly selected  training images and their down-sampled version. Early stage: mean over histogram of epochs one to four. Middle stage: mean over histogram of epochs 10 to 15. Final stage: mean over histogram of epochs 20 to 24. Top (before) and bottom (after) applying Eq. \ref{injectionM}.}
\label{fig:norm_analysis}
\vspace{-4mm}
\end{figure}

\subsection{Quality Aware Sample Injection}
\label{quaface}
To address the mentioned shortcomings of the classification framework, prevent Softmax centers from being distracted, and explore recognizable low-quality samples, we propose QAFace, a quality-aware injection procedure. Proposed method ignores unrecognizable samples and, at the same time, uses recognizable low-quality samples to add more valid uncertainty to the centers. The injection process is as follows:
\begin{equation}\label{injectionM}
 \small
 \begin{aligned}
	W_{y_i} = W_{y_i} + f(\widehat{||x_i||}) * \frac{x_i}{||x_i||},  
\end{aligned}
\end{equation}
where $\small{\widehat{||x_i||}}$ is the normalized version of feature magnitude. We normalize the feature magnitude via batch statistics: $\small{\mu}$ and $\small{\sigma}$. To relax $\small{\mu}$ and $\small{\sigma}$ from the batch size, we calculate them in an exponential moving average over the training iterations.
$\small{f(\widehat{||x_i||})}$ projects $\small{\widehat{||x_i||}}$ to have zero value for the features that have  $\small{\widehat{||x_i||}}$ lower than a threshold, $- \tau$, otherwise positive. 
\begin{equation}\label{sigmaEMA}
 \small
 \begin{aligned}
	\sigma_{t} = \alpha \sigma_t + (1-\alpha)\sigma_{t-1},  
\end{aligned}
\end{equation}

\begin{equation}\label{meanEMA}
 \small
 \begin{aligned}
	\mu_{t} = \alpha \mu_t + (1-\alpha)\mu_{t-1},  
\end{aligned}
\end{equation}

\begin{equation}\label{injectionM}
 \small
 \begin{aligned}
	\widehat{||x_i||} = \frac{||x_i|| - (\mu)}{\sigma},  
\end{aligned}
\end{equation}

\begin{equation}\label{injectionM_}
 \small
 \begin{aligned}
 f(\widehat{||x_i||})=\left\{ 
  \begin{array}{ c l }
    e^{-\widehat{||x_i||}} & \quad \textrm{if } \widehat{||x_i||} \geq -\tau, \\
    0                & \quad \textrm{else}.
  \end{array}
 \right.
 \end{aligned}
\end{equation}

Eqs.~\ref{injectionM} and \ref{injectionM_} together work in a way that 1) samples with $\widehat{||x||}$ lower than -$\tau$ would not affect the centers, 2) recognizable but low-quality samples will be emphasized during training, and 3) high-quality samples will receive less attention in comparison to recognizable low-quality samples. Using informative samples in metric-based FR training paradigm has been well-established \cite{schroff2015facenet}. Therefore, employing informative samples to add sample-wise comparison to the classification framework is of most importance. Our proposed algorithm adaptively:  1) assigns more weight to the recognizable low-quality (hard) samples, 2) ignores unrecognizable samples, and 3) puts less attention on easy high-quality samples.
Hence, our approach can be regarded as a kind of hard-sample mining, but without adding any computational burden on hard sample selection.
It is worth mentioning that, $\small{f(\widehat{||x_i||}) * \frac{x_i}{||x_i||}}$ is happening during the memorizing the representations in the memory. Consequently, we can re-write the Eq.~\ref{injectionM} as:
\begin{equation}\label{injectionMemory}
 \small
 \begin{aligned}
	W_{y_i} = W_{y_i} + M(\widehat{||x_i||}, x_i).  
\end{aligned}
\end{equation}
\begin{figure}[t]
\begin{center}
\includegraphics[width=1\linewidth]{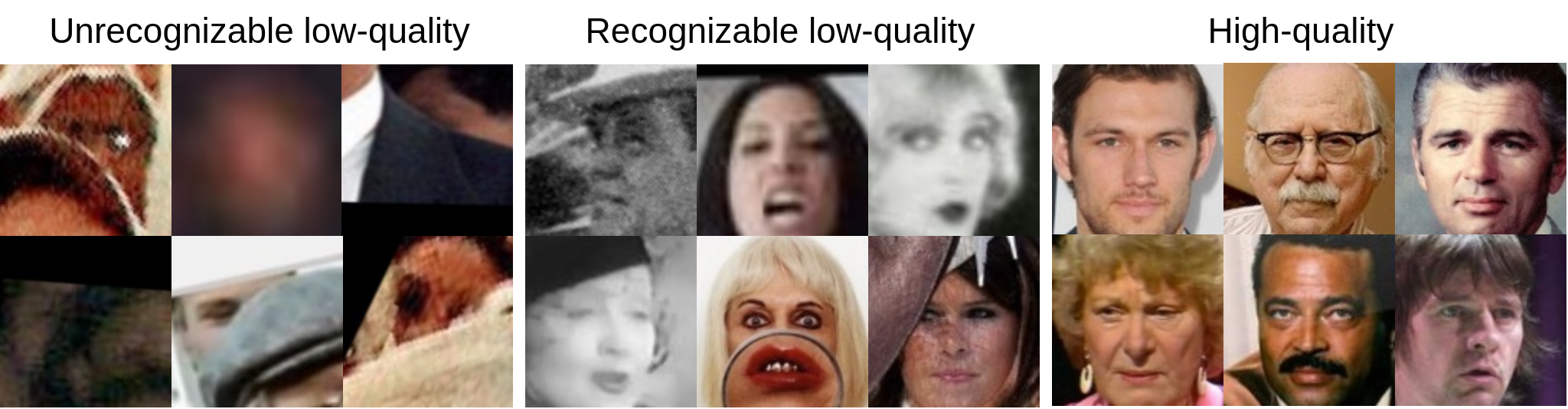}
\end{center}
\vspace{-3mm}
   \caption{Illustration of three types of samples with regard to the feature norm at the final stage of training. Left: samples that model ignored. Middle: Samples that are being emphasised. Right: Samples with high feature norm. }
\label{fig:typesOfSamples}

\end{figure}
\begin{table}[]
\addtolength{\tabcolsep}{-2.6pt} 
\small
\begin{center}
\caption{Performance (\%) of Arcface, VPL, and our method on the different down-sampled versions of LFW, CFP-FP, CALFW, CPLFW, AgeDB. 1:1 verification accuracy is reported.}
\begin{tabular}{m{1em}|c|ccccc}
\specialrule{.1em}{.1em}{.1em}\hline
          & Resolution   & LFW   & CFP-FP & CPLFW & CALFW & AgeDB \\ \hline
\multirow{3}{*}{\rotatebox{90}{ArcFace }} & 8$\times$8   & 71.86 & 56.92  & 56.43 & 57.56 & 54.48 \\
                         & 16$\times$16 & 96.60 & 84.21  & 82.76 & 84.30 & 78.20 \\
                         & original     & 99.83 & 98.27  & 92.08 & 95.45 & 98.28 \\ \hline
\multirow{3}{*}{\rotatebox{90}{VPL }}     & 8$\times$8   & 71.96 & 60.75  & 57.78 & 59.56 & 52.45 \\
                         & 16$\times$16 & 97.30 & 85.98  & 83.53 & 84.61 & 79.06 \\
                         & original     & 99.83 & 99.11  & 93.45 & 96.12 & 98.60 \\ \hline
\multirow{3}{*}{\rotatebox{90}{QAFace }}    & 8$\times$8   & 72.76 & 59.62  & 57.65 & 59.93 & 54.16 \\
                         & 16$\times$16 & 98.26 & 89.57  & 86.75 & 88.20 & 83.56 \\
                         & original     & 99.85 & 99.21  & 94.41 & 96.11 & 97.91 \\ 
\hline \specialrule{.1em}{.1em}{.1em}
                        
\end{tabular}
\label{table:lowresperformance}
\end{center}
\vspace{-10mm}
\end{table}

\begin{table*}[]
\small
\addtolength{\tabcolsep}{6pt} 
\begin{center}
\caption{Perfomance (\%) comparison of our method with other recent algorithms. 1:1 verification accuracy is reported on LFW, CFP-FP, CPLFW, AgeDB.}
\begin{tabular}{cc|ccccccc}
\specialrule{.1em}{.1em}{.1em}\hline
\multicolumn{1}{c|}{\multirow{2}{*}{Method}} & \multirow{2}{*}{Venue} & \multicolumn{5}{c}{Verification accuracy}                   & \multicolumn{2}{c}{TAR@FAR=$1e-4$} \\ \cline{3-9} 
\multicolumn{1}{c|}{}                        &                        & LFW   & CFP-FP & CPLFW & CALFW & \multicolumn{1}{c|}{AgeDB} & IJB-B             & IJB-C             \\ \hline
\multicolumn{1}{c|}{Wang {\it et al.} \cite{wang2018cosface}}                 & CVPR18                 & 99.81 & 98.12  & 92.28 & 95.76 & \multicolumn{1}{c|}{98.11} & 94.80             & 96.37             \\
\multicolumn{1}{c|}{Deng {\it et al.} \cite{deng2019arcface}}                 & CVPR19                 & 99.83 & 98.27  & 92.08 & 95.45 & \multicolumn{1}{c|}{98.28} & 94.25             & 96.03             \\
\multicolumn{1}{c|}{Sun {\it et al.} \cite{sun2020circle}}              & CVPR20                 & 99.73 & 96.02  &       & -     & \multicolumn{1}{c|}{-}     & -                 & 93.95             \\
\multicolumn{1}{c|}{Deng {\it et al.} \cite{deng2020sub}}      & ECCV20                 & 99.80 & 98.80  &       &       & \multicolumn{1}{c|}{98.31} & 94.94             & 96.28             \\
\multicolumn{1}{c|}{Wang {\it et al.} \cite{wang2020mis}}              & AAAI20                 & 99.80 & 98.28  & 92.83 & 97.95 & \multicolumn{1}{c|}{96.10} & 93.6              & 95.2              \\
\multicolumn{1}{c|}{Huang {\it et al.} \cite{huang2020curricularface}}          & CVPR20                 & 99.80 & 98.37  & 93.13 & 96.20 & \multicolumn{1}{c|}{98.32} & 94.8              & 96.1              \\
\multicolumn{1}{c|}{Kim {\it et al.} \cite{kim2020broadface}}               & ECCV20                 & 99.85 & 98.63  & 93.17 & 96.20 & \multicolumn{1}{c|}{98.38} & 94.97             & 96.38             \\
\multicolumn{1}{c|}{Shi {\it et al.} \cite{shi2020towards}}                  & CVPR20                 & 99.78 & 98.64  & -     & -     & \multicolumn{1}{c|}{-}     & -                 & 96.6              \\
\multicolumn{1}{c|}{Kim {\it et al.} \cite{kim2020groupface}}               & CVPR20                 & 99.85 & 98.63  & 93.17 & 96.20 & \multicolumn{1}{c|}{98.28} & 94.93             & 96.26             \\
\multicolumn{1}{c|}{Chang {\it et al.} \cite{chang2020data}}                     & CVPR20                 & 99.83 & 98.78  & -     & -     & \multicolumn{1}{c|}{-}     & -                 & 94.61             \\
\multicolumn{1}{c|}{Meng {\it et al.} \cite{meng2021magface}}                 & CVPR21                 & 99.83 & 98.46  & 92.87 & 96.15 & \multicolumn{1}{c|}{98.17} & 94.51             & 95.97             \\
\multicolumn{1}{c|}{Deng {\it et al.} \cite{deng2021variational}}                     & CVPR21                 & 99.83 & 99.11  & 93.45 & 96.12 & \multicolumn{1}{c|}{98.60} & 95.56             & 96.76             \\ \hline
\multicolumn{1}{c}{QAFace} &                                             & \textbf{99.85}     &     \textbf{99.21}  &   \textbf{94.41}   &   {96.11}   & \multicolumn{1}{c|}{97.91}     & \textbf{95.67}                 & \textbf{97.20}                 \\ 
\hline \specialrule{.1em}{.1em}{.1em}
\end{tabular}
\label{table:abc}
\vspace{-5mm}
\end{center}
\end{table*}

\subsection{Distractor Samples}
The major advantage of the QAFace over \cite{deng2021variational} is the ability to identify the unrecognizable from recognizable samples and emphasize the recognizable low-quality samples in the injection process. To this end, we employ the magnitude of the feature vector as a proxy for the input sample recognizability \cite{meng2021magface}. We perform an experiment to demonstrate how the feature magnitude is affected by recognizability and how our method can use hard samples to reduce the gap between the representation of low and high-quality samples.
We randomly select a subset of 10K images of the training data and down-sampled them to the different levels ($8\times 8$, and $16 \times 16$). Then, at the end of every epoch, we save the magnitude of the representations obtained from original and down-sampled images, see Fig. \ref{fig:norm_analysis}. From the results shown in Fig.~\ref{fig:norm_analysis} (top) we can observe that as the recognizability increases, the magnitude of representations increases as well, i.e.,\ $green > blue > orange$. 

Complete overlapping of the distribution of down-sampled with the original instances is an ideal scenario, which means that the model became quality-agnostic. The model progressively learns to increase the lower bond of feature magnitude to narrow the gap between original and low-quality samples. At the early stages of the training, the full range of feature magnitude is around 25 (from 5 to 30). Then in the final stages, the range narrows to 15 (from 15 to 30). Also, the mean of the original image distribution (green) is always around 23, which shows that our proposed method can effectively involve low-quality samples in training without reducing the performance on the high-quality samples. 
%
%
%
%
%

Another observation from Fig. \ref{fig:norm_analysis} (bottom) is the necessity of normalizing the magnitude. Without applying Eq. \ref{injectionM} on the feature magnitude, the threshold for ignoring unrecognizable samples changes as the training progresses. Eq. \ref{injectionM} omits the bias from the distribution of features magnitude caused by the training stage. Therefore, we can choose a fixed $\tau$ for ignoring the unrecognizable and emphasizing the hard samples. 


\subsection{Complement to Angular-Margin Gradient}
\begin{figure}[t]
\begin{center}
\includegraphics[width=1\linewidth]{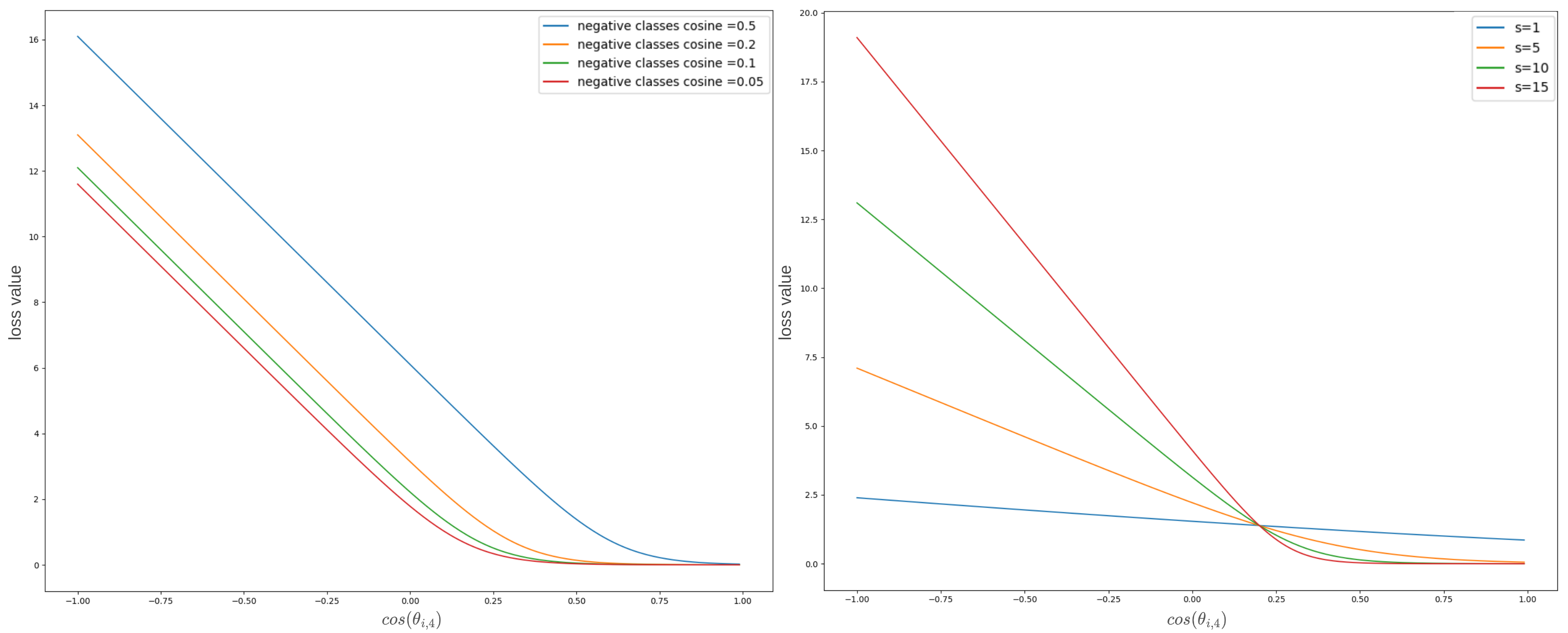}
\end{center}
\vspace{-3mm}
   \caption{Curves of $\small{p_{y_i} = \frac{e^{s(cos(\theta_{y_i}))}}{{\sum_{{j=1 }}^{C}{e^{s (cos(\theta_j))}}}}}$, when $y_i=4$ and $cos(\theta_{y_i})$ changes from -1 to 1.}
\label{fig:sandp}
\end{figure}

Unlike triplet and contrastive losses, softmax-based losses are not subject to explicit easy/hard sample mining \cite{wen2021sphereface2,nourelahi2022explainable}. In this section, using a simple toy example, we show that the Softmax-based losses implicitly benefit easy/hard sample mining by their gradient. Additionally, we elaborate on the ability of the proposed $\small{f(\widehat{||x_i||})}$ to complement the Softmax learning signal (gradient).
Consider a four-identity classification. For a given sample $x_i$ with ground truth identity $\small{y_i=4}$, the logits are $\small{{cos(\theta_{1}), cos(\theta_{2}), cos(\theta_{3}), cos(\theta_{4})}}$. In Fig. \ref{fig:sandp}, we plot the loss value for a fixed $cos(\theta_{j}),j\neq 4$ while $cos(\theta_{y_i=4})$ changes from -1 to 1. The first observation from Fig. \ref{fig:sandp} (right) is that the scaling parameter $s$ is tuning the sensitivity of the loss function. As the scaling value increases, the slope of the loss function (gradient) increases \cite{zhang2019adacos}.

Moreover, $s$ directly influences the point that samples would be recognized as easy. Easy samples would barely experience change, i.e.,\ low slope, while hard samples receive high gradient value, i.e.,\ high slope. 
Also, in Fig. \ref{fig:sandp} (left), we demonstrate that the drawback here is the monotonicity of the gradient among the hard samples. In other words, some samples are unrecognizable; however, their gradient is equivalent to those of informative but hard samples. Consequently, the model tries to overfit the unrecognizable samples because there is no identity information on those instances \cite{shi2020towards}. Our proposed method tries to compensate for this effect by ignoring the unrecognizable samples during the injection. It does not further involve the unrecognizable samples in the injection process and ignores them using the proposed feature weighting paradigm. As a result, our method justifies the classifier direction to tolerate more variation toward hard and informative samples and plays a complementary role to the Softmax-based learning signal.

\section{Experiments}
\subsection{Datasets}
We employ
Webface4M \cite{zhu2021webface260m} as our training data, which contain 
4 million samples of around 200,000 identities, Table \ref{table:abc}. For evaluation of our method, we use CFP-FP \cite{sengupta2016frontal}, CPLFW \cite{zheng2018cross}, CALFW \cite{zheng2017cross}, LFW \cite{huang2008labeled}, AgeDB \cite{moschoglou2017agedb}, IJB-B \cite{whitelam2017iarpa}, and IJB-C \cite{maze2018iarpa}.
Based on the datasets evaluation protocols, we report 1:1 verification accuracy for  CFP-FP, LFW, CPLFW, CALFW, and AgeDB datasets.
For IJB-B \cite{whitelam2017iarpa} and IJB-C \cite{maze2018iarpa}, we report the True Acceptance Rate (TAR) over the False Acceptance Rate of $1e-4$.
\begin{figure}[t]
\begin{center}
\includegraphics[width=0.95\linewidth]{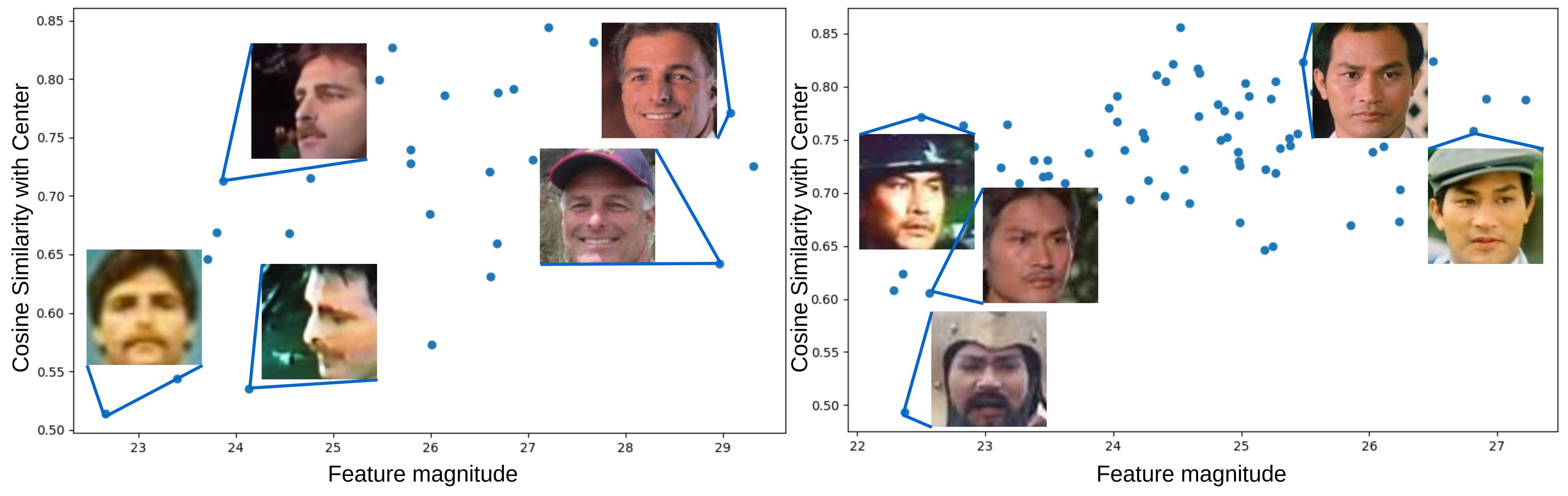}
\vspace{-5mm}
\end{center}
   \caption{Illustration of sample-to-center cosine similarity for two randomly selected subjects. High-norm samples are very similar to the class center. Low-norm samples have lower similarity with Softmax center.}
\label{fig:long}
\vspace{-5mm}
\label{fig:SimVsCls}
\end{figure}

\subsection{Training Settings}
We use \cite{deng2020retinaface} to detect five landmarks in each image. Then images are aligned and rescaled to $112\times 112$, following the setting in \cite{deng2019arcface}. We adopt ResNet \cite{deng2019arcface} for the backbone. The model is trained for 24 epochs with Arcface loss. The optimizer is SGD, with the learning rate starting from 0.1, which is decreased by a factor of 10 at epochs \{10, 16, 22\}. The optimizer weight-decay is set to 0.0001, and the momentum is 0.9. During training, the mini-batch size on each GPU is 512, and the model is trained using two Quadro RTX 8000. Following \cite{he2020momentum}, $\gamma$ in Fig. \ref{fig:short} is 0.99. In calculating the $\mu$ and $\sigma$, $\alpha$ in Eq. \ref{sigmaEMA} and \ref{meanEMA}, is 0.99. Given a pair of images, the cosine distance between the representations is the metric during inference. 

\begin{figure}[t]
\begin{center}
\includegraphics[width=0.85\linewidth]{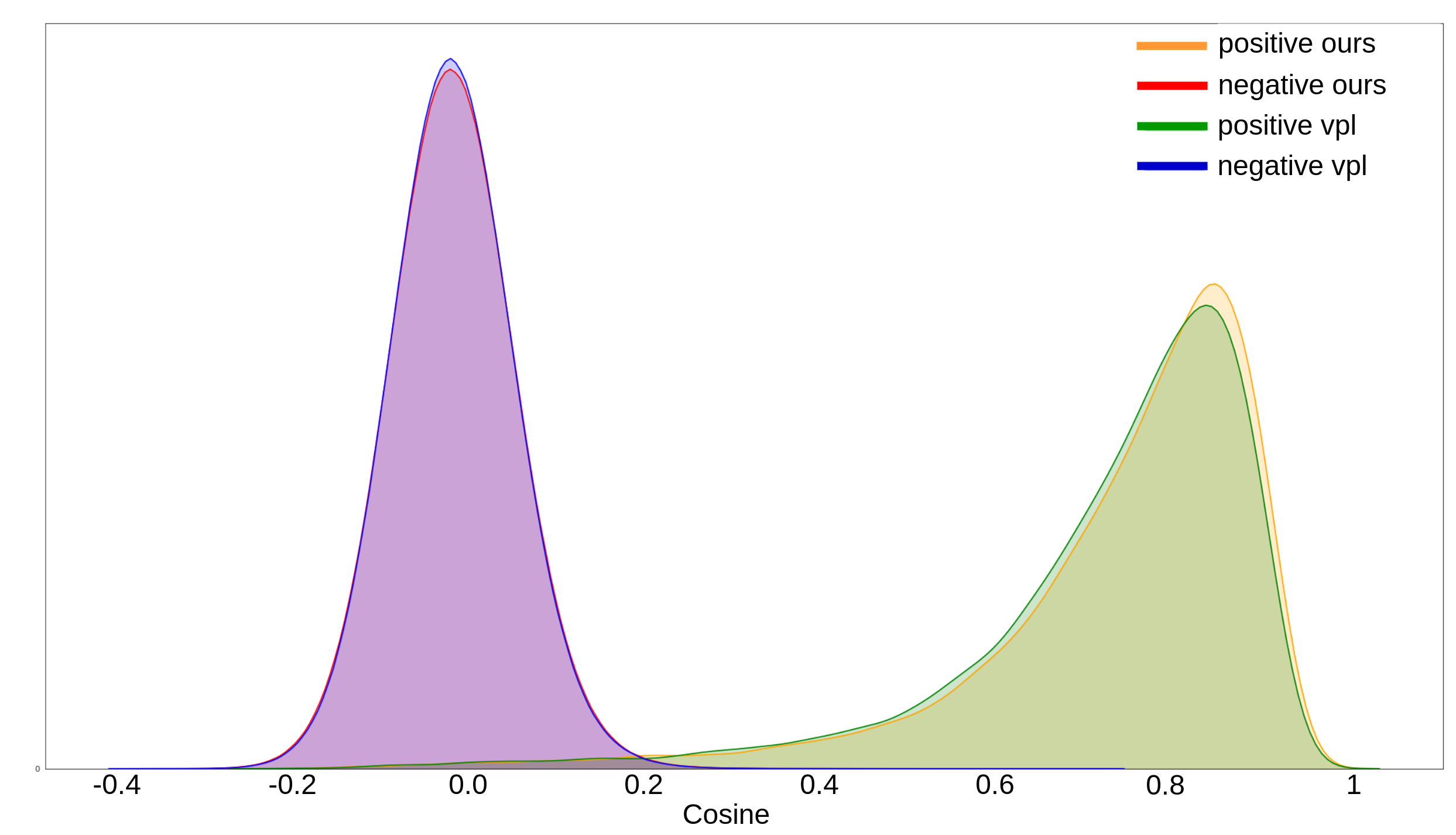}
\vspace{-5mm}
\end{center}
   \caption{Comparison between pair-wise similarity score on the IJB-C dataset obtained from VPL and QAFace.}
\label{fig:long}
\vspace{-2mm}
\label{fig:SimVsCls}
\end{figure}

\subsection{Ablation Study}
\subsubsection{Impact of the Memory Length}
In \cite{he2020momentum}, the memory is a dynamic queue of representations. The whole memory has a length of $|M|$, and queuing new samples results in de-queuing the oldest samples. Here the length of memory is equal to the number of classes. Therefore, we should memorize the last iteration in that every instance in the memory was updated. In this way, we can prevent from employing outdated representations in the injection. For instance, if the training is on the iteration $I$ and a specific instance in memory was updated on $I-\Delta t$, if $\Delta t$ is larger than a threshold, that specific memory instance cannot be used during the injection. It is shown that in the early epochs, the changes in the feature space are drastic; after that, it is negligible \cite{deng2021variational}. Therefore, we start the injection after the fourth epoch of training. Table \ref{table:ablationDeltat} shows the ablation experiments on $\Delta t$. In these experiments, we fixed  $\tau$  to 2. 
We increase  $\Delta t$ with the interval of 500 iterations. As illustrated in Table \ref{table:ablationDeltat}, the performance constantly improves from $\Delta t=0$ to $\Delta t=1000$, and after that, it starts to degrade.  

\begin{table}[]
\addtolength{\tabcolsep}{1pt} 
\small
\begin{center}
\caption{Ablation of $\Delta t$. The metrics are the same as Table \ref{table:abc}.}
\vspace{-3mm}
\begin{tabular}{c|ccc|cc}
\specialrule{.1em}{.1em}{.1em}\hline
\multirow{2}{*}{$\Delta t$} & \multicolumn{3}{c|}{Verification Accuracy} & \multicolumn{2}{c}{TAR@FAR:$1e-4$} \\ \cline{2-6} 
                            & LFW          & CFP-FP        & CPLFW       & IJB-B           & IJB-C           \\ \hline
0                           & 99.71        & 98.40         & 92.01       & 95.26           & 96.4            \\
500                         & 99.80        & 98.81         & 92.84       & 95.46           & 96.75           \\
1000                        & 99.85        & 99.21         & 94.41       & 95.67           & 97.20           \\
1500                        & 99.78        & 99.01         & 93.12       & 95.45           & 96.87           \\
2000                        & 99.69        & 98.33         & 92.95       & 95.14           & 96.35           \\ 
\hline \specialrule{.1em}{.1em}{.1em}
\end{tabular}
\label{table:ablationDeltat}
\vspace{-7mm}
\end{center}
\end{table}

\subsubsection{Impact of Threshold ($\tau$)}
We fix the memory length to 1000 iterations. Then we investigate different values for the threshold ($\tau$) in Eq.~\ref{injectionM}. As illustrated in Table \ref{table:ablationTau}, the performance on IJB-B and IJB-C constantly increases with changing  $\tau$ from zero to 2. At $\tau=0$, only samples with $\widehat{||x||}$ above the zero are involved in the injection. Consequently, the model performance decreases when the input comes from datasets like IJB-B and IJB-C, which contain low-quality samples \cite{shi2020towards}. On the other hand, the results in clean datasets like CFP, CPLFW,  and LFW are reasonably good \cite{shi2020towards}.

\begin{table}[]
\addtolength{\tabcolsep}{3pt} 
\small
\begin{center}
\caption{Ablation of $\tau$. The metrics are the same as Table \ref{table:abc}.}
\vspace{-2mm}
\begin{tabular}{c|ccc|cc}
\specialrule{.1em}{.1em}{.1em}\hline
\multirow{2}{*}{$\tau$} & \multicolumn{3}{c|}{Verification Accuracy} & \multicolumn{2}{c}{TAR@FAR:1e-5} \\ \cline{2-6} 
                        & LFW           & CFP-FP       & CPLFW       & IJB-B           & IJB-C           \\ \hline
0                       & 99.86         & 99.23        & 93.20       & 95.03           & 96.12           \\
1                       & 99..83        & 99.11        & 93.14       & 95.41           & 96.91           \\
2                       & 99.85         & 99.21        & 94.41       & 95.67           & 97.20           \\
3                       & 99.83         & 99.02        & 93.02       & 95.51           & 96.84           \\
4                       & 98.76         & 98.45        & 92.32       & 95.02           & 96.20           \\ 
\hline \specialrule{.1em}{.1em}{.1em}
\end{tabular}
\label{table:ablationTau}
\vspace{-7mm}
\end{center}
\end{table}

\subsection{Impact of Augmentation}
For data augmentations, we used random cropping and down-sampling \cite{joshi2021fdeblur, zafari2022attention}. On-the-fly data augmentation provides more diverse training data. However, as illustrated in  Fig. \ref{fig:aug}, it increases the occurrence of unrecognizable samples. We perform experiments on our method with and without the presence of data augmentation. Accordingly, we can show that our method can effectively ignore unrecognizable samples and, at the same time, benefits from more training instances, see Fig. \ref{fig:typesOfSamples}. As shown in Table \ref{table:augmentation}, the model gains performance on the IJB-B and IJB-C datasets by increasing the probability of augmentations. As these datasets contain low-quality samples, down-sampling leads to more performance improvement than random cropping.   
\begin{table}[]
\small
\caption{Ablation of augmentation probability, TAR@FAR=1e-4.}
\vspace{-5mm}
\begin{center}
\begin{tabular}{ccc|cc}
\specialrule{.1em}{.1em}{.1em}\hline
probability & cropping                  & down-sampling             & IJB-B & IJB-C  \\ \hline
0.0         & -                         & -                         & 95.34 & 96.60 \\ \hline
0.1         & -                         & \checkmark & 95.56 & 96.89 \\ \hline
0.1         & \checkmark & -                         & 95.41 & 96.65 \\ \hline
0.1         & \checkmark & \checkmark & 95.57 & 96.95 \\ \hline
0.2         & -                         & \checkmark & 95.63  & 96.91 \\ \hline
0.2         & \checkmark & -                         & 95.45 & 96.67 \\ \hline
0.2         & \checkmark & \checkmark & 95.67 & 97.20 \\ 
\hline \specialrule{.1em}{.1em}{.1em}
\end{tabular}
\label{table:augmentation}
\vspace{-7mm}
\end{center}
\end{table}

\subsection{Comparison with state-of-the-art }
Table \ref{table:abc} shows the proposed method's performance compared to the state-of-the-art algorithms. For better clarification, we explain our observation in two parts. For the results on LFW, CPLFW, CALFW, CFP-FP, and AgeDB, it is essential to mention that QAFace is built upon putting more emphasis on the low-quality samples and making these samples' representation more similar to the high-quality samples' features. Consequently, the performance gain in these datasets is marginal, as they contain almost high-quality samples \cite{shi2020towards}. Although the performance is saturated in most of these datasets, our method strives to increase the 1:1 verification accuracy for the CFP-FP and CPLFW datasets.
The IJB-B and IJB-C datasets are more challenging and have images/frames with diverse quality.
Results on the IJB-B and IJB-C datasets show the superiority of our approach in more general face recognition. As these datasets contain low-quality and high-quality images, the performance gain in these datasets is more evident. In IJB-B, compared to VPL, QAFace improves the TAR at FAR=$1e-4$. 


\section{Conclusion}
This work argues the importance of integrating sample-wise similarity to the Softmax framework. Also, we showed that existing angular-margin-based loss functions could be distracted by the unrecognizable samples in the dataset. Inspired by the well-established idea of hard sample mining in the sample-to-sample comparison framework, we proposed a weighting scenario to ignore unrecognizable samples and emphasize recognizable low-quality samples during the injection. We empirically showed the effect of ignoring unrecognizable samples by improving the similarity score between positive samples in the IJB-C dataset. Also, We analyzed the proposed function for weighting. Our proposed approach is based on the simple idea of using the norm of features as the proxy for the recognizability of face images.
Furthermore, we empirically showed the effect of the quality of face images on the magnitude of features. We demonstrated that there is a direct proportion between the face image quality and the magnitude of its representation. Our approach could successfully outperform all of its competitors in five out of seven evaluation benchmarks, including the IJB-B and IJB-C datasets. 

\section{Acknowledgement}
This research is based upon work supported by the Office of the Director of National Intelligence (ODNI), Intelligence Advanced Research Projects Activity (IARPA), via IARPA R\&D Contract No. 2022-21102100001. The views and conclusions contained herein are those of the authors and should not be interpreted as necessarily representing the official policies or endorsements, either expressed or implied, of the ODNI, IARPA, or the U.S. Government. The U.S. Government is authorized to reproduce and distribute reprints for Governmental purposes notwithstanding any copyright annotation thereon.
{\small
\bibliographystyle{ieee_fullname}
\bibliography{egbib}
}

\end{document}